\documentclass[letterpaper, 10 pt, conference]{ieeeconf}

\usepackage{listings}
\usepackage{algpseudocode}
\usepackage{graphicx}
\usepackage{subfig}
\usepackage{color}
\usepackage{hyperref}
\usepackage{tabularx}
\usepackage{booktabs,tabularx}
\usepackage{balance}
\usepackage{nth}
\usepackage{pgf-pie}
\usepackage{cite}
\usepackage{float}

\definecolor{dkgreen}{rgb}{0,0.6,0}
\definecolor{gray}{rgb}{0.5,0.5,0.5}
\definecolor{mauve}{rgb}{0.58,0,0.82}

\newlength{\tempheight}
\newlength{\tempwidth}

\newcommand{\rowname}[1]
{\rotatebox{90}{\makebox[\tempheight][c]{\textbf{#1}}}}

\newcommand{\columnname}[1]
{\makebox[\tempwidth][c]{\textbf{#1}}}

\title{\LARGE \bf Autonomous Soil Collection in Environments With Heterogeneous Terrain
}

\author{Andrew Dudash$^{1}$, Beyonce Andrades, Ryan Rubel, Mohammad Goli, Nathan Clark, William Ewald %
\thanks{$^{1}$Andrew Dudash
        Noblis, 2002 Edmund Halley Drive, Reston, VA 20191
        {\tt\small andrew.dudash@noblis.org}}%
}

\IEEEoverridecommandlockouts

\begin{document}

\thispagestyle{empty}
\pagestyle{empty}


\maketitle

\begin{abstract}
To autonomously collect soil in uncultivated terrain, robotic arms must distinguish between different amorphous materials and submerge themselves into the correct material. We develop a prototype that collects soil in heterogeneous terrain. If mounted to a mobile robot, it can be used to perform soil collection and analysis without human intervention. Unique among soil sampling robots, we use a general-purpose robotic arm rather than a soil core sampler.
\end{abstract}

\section{INTRODUCTION}

\begin{figure}[H]
  \centering
  \subfloat[The objects are shaped and homogeneous.]{
    \label{fig:homo-object}
    \includegraphics[keepaspectratio,width=0.40\linewidth]{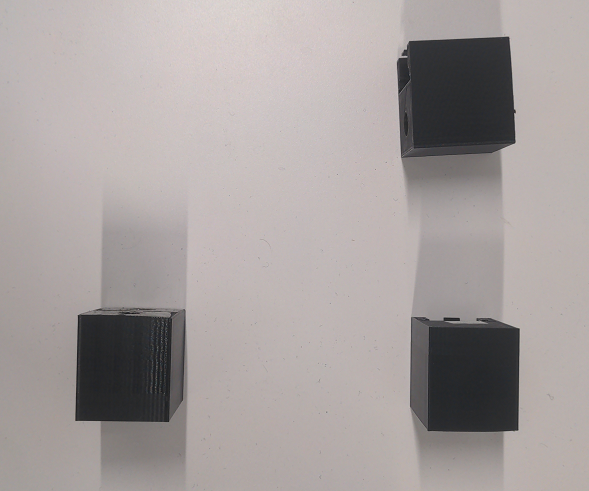}
  }\hspace{0.01em}
  \subfloat[The material is amorphous and homogeneous.]{
    \label{fig:homo-material}
    \includegraphics[keepaspectratio,width=0.40\linewidth]{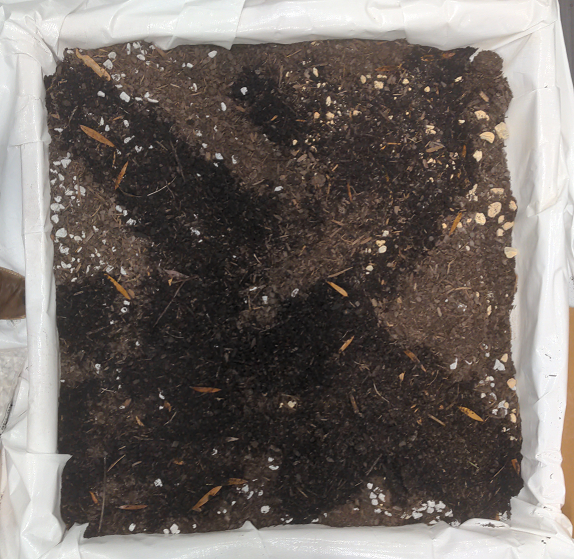}
  }

  \subfloat[The objects are shaped and heterogeneous.]{
    \label{fig:hetero-object}
    \includegraphics[keepaspectratio,width=0.40\linewidth]{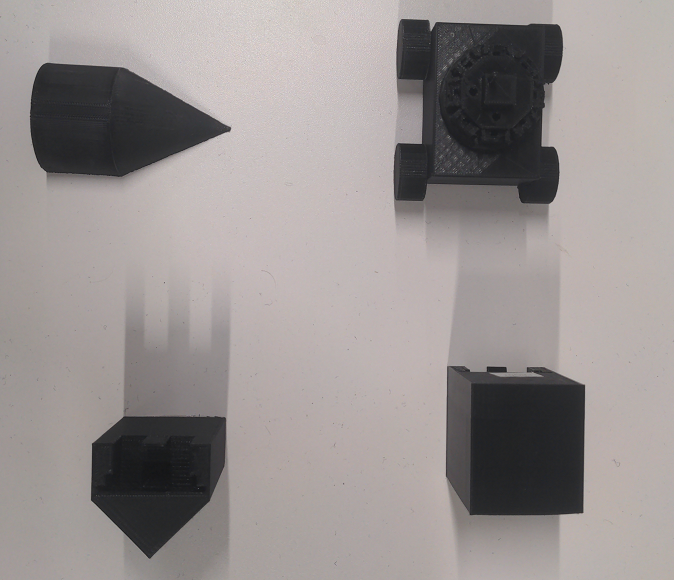}
  }
  \hspace{0.01em}
  \subfloat[The material is amorphous and heterogeneous.]{
    \label{fig:hetero-material}
    \includegraphics[keepaspectratio,width=0.40\linewidth]{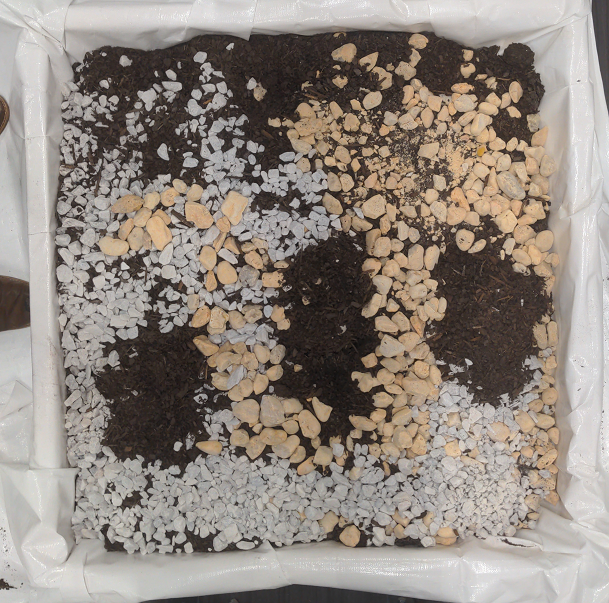}
  }
  \caption{The prototype described in this system discovers and collects soil samples in the situation shown in Figure \ref{fig:hetero-material}.}
  \label{figure:motivation1}
\end{figure}

Consider the objects in Figure \ref{fig:homo-object}. A robotic pick-and-place system, like a sorting robot, can use a computer vision system to orient an end effector, like a suction cup or gripper, over the object of interest and pick it up. In this case, there is a single shaped object type to pick up. In contrast, the mulch in Figure \ref{fig:homo-material} is not a shaped object. The ground is an amorphous surface of dirt. To pick the dirt, the arm must submerge its gripper within the material. Detection is simpler because the entire surface is pickable, but positioning is made more difficult because the object must partially submerge itself into the material. In Figure \ref{fig:hetero-object}, the objects are again shaped, but now there are multiple types of objects. If we only want to pick a certain type of object, we must have a way to distinguish the different types. We have extended the easiest sampling case, shaped with homogeneous objects, to two new problems: amorphous but homogeneous and shaped but heterogeneous.

In the final figure, Figure \ref{fig:hetero-material}, the arm is again tasked to pick an amorphous material by partially submerging itself, but this time the surface is heterogeneous; the amorphous and heterogeneous cases are combined. The robot needs to pick soil while avoiding rocks.

\begin{figure}[H]
  \centering
  \includegraphics[keepaspectratio,width=0.6\linewidth]{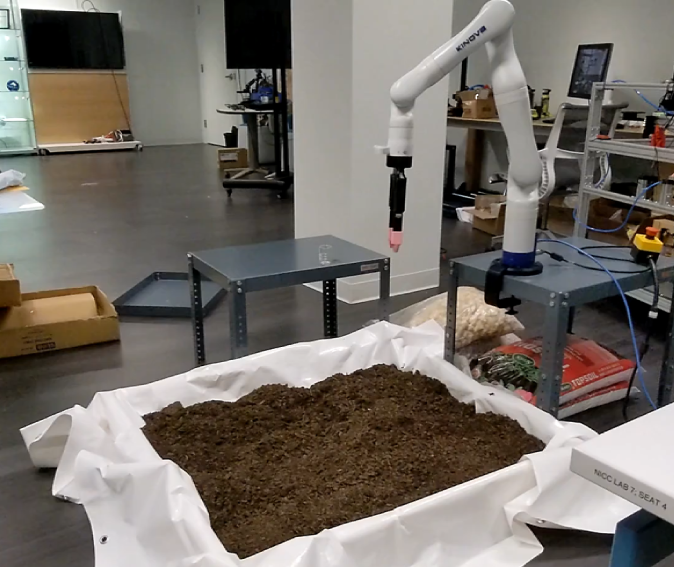}
  \caption{Our prototype uses an arm with 7-DOF, a wrist mounted camera, and a two finger end effector.}
  \label{fig:prototype1} 
\end{figure}

In this paper, we develop a prototype, shown in Figure \ref{fig:prototype1}, a control system capable of picking a specific material from a heterogeneous surface---the problem in Figure \ref{fig:hetero-material}---and test it on a physical surface. We list our specific contributions:

\begin{itemize}
\item We build a robotic arm control system that can autonomously retrieve soil samples in heterogeneous amorphous terrain. We validate the prototype's design and measure its performance experimentally.
\item We build a computer vision module that can automatically distinguish soil areas from other material. We measure its performance on a test dataset and in experiment, and evaluate its design in an ablation study.  
\end{itemize}

\section{RELATED WORK} \label{section:related}

Robotic object picking has a variety of applications: weeding\cite{Kurtser_Lowry_2023}, soil collection \cite{Olmedo_Barczyk_Lipsett_2020}, plant disease detection \cite{Kurtser_Lowry_2023}, cotton harvesting\cite{Gharakhani_Thomasson_Lu_2022}, package sorting \cite{package_sorting}, manufacturing related gripping \cite{metal_gripping}, and mail delivery \cite{mail_delivery}.

Picking amorphous materials, like soil, has been researched. In 2022, Paul et al. built a robotic food handler that picks minced onions from a container and garnishes food trays\cite{Paul2022AR2}. Their application is similar to our homogeneous amorphous material case in Figure \ref{fig:homo-material}, but with a controlled environment. Our system is instead focused on the heterogeneous amorphous case, Figure \ref{fig:hetero-material}.

Commercial autonomous soil analysis vendors use extraction rods that impale soil and then retrieve a small sample\footnote{Robo Ag and Godelius both use extraction rods.}. This is effective for agriculture, but requires niche dedicated hardware and must avoid coring rocks, concrete or other obstructions. We suggest an alternative solution using a two finger end effector. By using a modular robotic arm with a changeable end effector, specialized hardware is not needed for our system.

Our computer vision approach is similar to existing approaches in robotic agriculture research. Prior research, by Taconet et al.\cite{Taconet_Vannier_Le_Hegarat-Mascle_2010}, uses computer vision and semantic segmentation for soil analysis and determining soil properties, such as clod size\cite{Azizi_Abbaspour-Gilandeh_Vannier_Dusseaux_Mseri-Gundoshmian_Moghaddam_2020}, after collection rather than distinguishing soil from other terrain during collection. In 2018, Milioto, Lottes, and Stachniss developed a prototype that uses a deep learning based semantic segmentation model---the same technique in this paper---to detect and localize plants and weeds, but their prototype does not collect any crops or weeds\cite{8460962}. Semantic segmentation of weeds and crops was previously done by Champ et al. in 2020\cite{Champ2020InstanceSF}, Sheikh et al. in 2020 \cite{Sheikh2020GradientAL}, and Baravdish and Ranawaka in 2023\cite{Baravdish2023SemanticSO}. In particular, Baravdish and Ranwaka use a U-Net based architecture, like this paper, and Sheikh et al. use transfer learning, similar to the auxiliary task learning used in this paper.

Although there is significant work in robotic soil sampling and semantic segmentation for robotic agriculture, we were unable to find a physical prototype that combined vision and sample collection to handle heterogeneous amorphous material.


\section{ARCHITECTURE} \label{section:methodology}


Our prototype has three parts: the robot hardware, the computer vision system software, and the sample control system. The sensor hardware provides images to the computer vision system which  provides positions to the sample control system which forwards the positions to a motion planner, and then the motion planner controls the arm hardware.

\subsection*{Hardware}

The robotic hardware is an arm with a vision sensor and a single board computer. The arm is a Kinova Gen3 arm with 7 degrees of freedom and a wrist mounted Intel RealSense\textsuperscript{TM} Depth Camera and 2F-140 Robotiq gripper. The sensor is an integrated infrared projector and stereo vision camera. The gripper is a two finger gripper. To improve soil retention, we designed, 3D printed, and attached fluted extensions to each of the gripper fingers. The extensions contain a small gap for holding soil. Despite this modification, the gripper can still be used for other grasping tasks. The single board computer that controls the arm is an Nvidia Jetson Orin running GNU/Linux and Robot Operating System (ROS).

\subsection*{Computer Vision System}
In our system, all soil can be sampled. Rocks, trees, grass, and roots are unpickable and must be avoided.

\begin{figure}
  \centering
  \subfloat[The wrist camera provides an image of the environment.]{
    \label{fig:raw1}
    \includegraphics[keepaspectratio,width=0.45\linewidth]{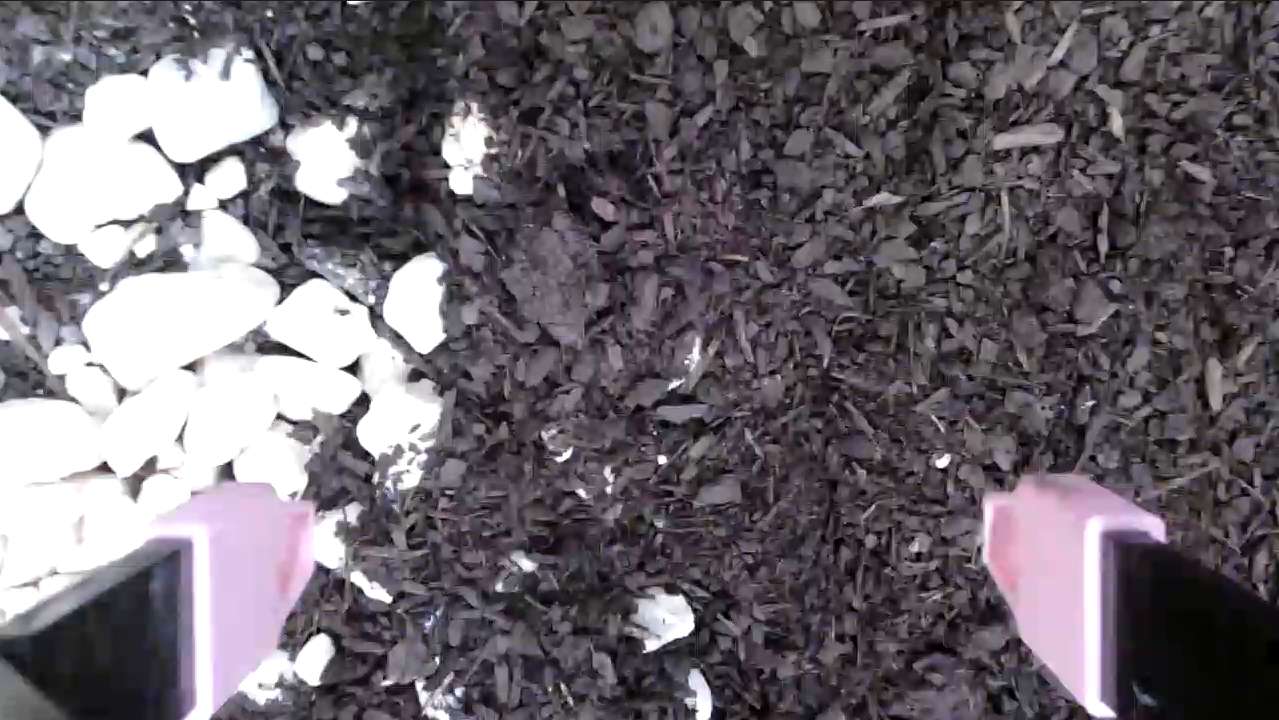}
  }

  \subfloat[The image is segmented into pickable and unpickable.]{
    \label{fig:segmentation1}
    \includegraphics[keepaspectratio,width=0.45\linewidth]{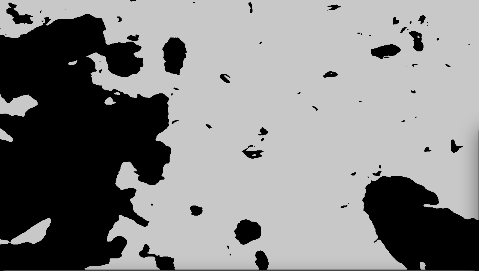}
  }\hspace{0.01em}
  \subfloat[Contour detection is used to group different regions of the image.] {
    \label{fig:contour1}
    \includegraphics[keepaspectratio,width=0.45\linewidth]{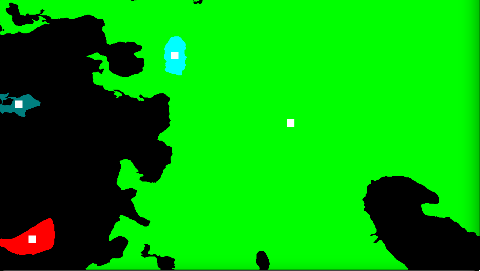}
  }

  \caption{We use a segmentation model to classify each pixel of the image and then we use a border following algorithm to break the segmented image into groups of pickable objects.}
  \label{fig:cv-pipeline1}
\end{figure}

To do this, the computer vision system converts a camera image into a target position in three steps: semantic segmentation, contour detection, and then a coordinate transform. First, the computer vision system segments the camera image into pickable and unpickable pixels. The vision system then generates contours of different pickable regions using a border following algorithm\cite{Suzuki1985TopologicalSA}. The centers of the contours for each pickable region are the target positions. Figure \ref{fig:cv-pipeline1} shows a camera image segmented and then turned into a series of image coordinates. Once the image coordinates are generated, they are transformed into 3D space by a transform with the camera matrix, and then, by another transform, into the robotic arm's base coordinate frame. The points are passed to the sample control system.

The computer vision model is a U-Net encoder-decoder architecture\cite{unet1} pretrained on a surrogate task: colorization. U-Net is a standard semantic segmentation model. It uses skip connections between layers of equal size to improve performance by sharing context between the encoder and decoder layers.

\begin{figure}
  \label{fig:cv-results}
  \centering
  \subfloat{
    \label{fig:cv-results-row1}
    \includegraphics[keepaspectratio, width=85mm]{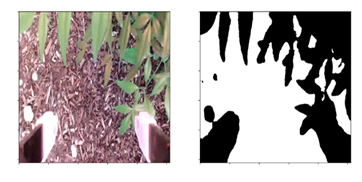}
  }

  \subfloat{
    \label{fig:cv-results-row2}
    \includegraphics[keepaspectratio, width=85mm]{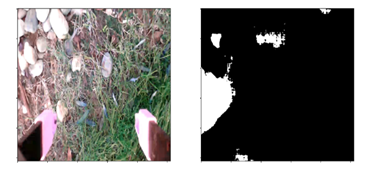}
  }

  \caption{The output of image segmentation is shown for two test images. The first column is the RGB images, the input, and the right column is the predicted image segmentation mask where white pixels are pickable areas, soil, and black pixels are unpickable areas.}
  \label{fig:cv-result1} 
\end{figure}

During development, we noticed that our pickable and unpickable regions were each associated with one color but that the intensity of the color might change under different lighting conditions, so we pretrained the model on colorization to improve robustness. To pretrain on color, we modified the model to predict RGB values and trained it on greyscale versions of our training data. We used the weights trained on this auxiliary task. The final model outputs a binary image. Example input and output of the final model is shown in Figure \ref{fig:cv-result1}.

The training data set is small. We created our own dataset to train our model. We collected 50 images of a makeshift dirt bed with rocks and 100 outdoor images of a landscaped office exterior. The latter images included trees, bushes, and mulch piles. To capture the images, the robotic arm was placed at random depths with the camera pointed down. These collected images were manually labeled using LabelStudio. The labelled training data was a set of images and image masks showing pickable regions.

Next, all images in the dataset were scaled to a size of 512x512 pixels. This simplified the computer vision model by allowing a stride length of 2 to be used for all convolution layers.

Because the dataset was small, we used data augmentation to increase the size of the training set: 150 hand labeled images.  On each batch, image rotation and flipping were applied randomly. When the model had finished training, at 200 epochs, each with a batch size of 4 images, it had used a total of 5,000 augmented images.

The dataset was shuffled and split into a training, validation, and test set.
\begin{itemize}
\item 70\% of the set was training.
\item 20\% of the set was validation.
\item 10\% of the set was testing.
\end{itemize}

The U-Net model was trained with the following hyperparameters:
\begin{itemize}
\item Batch Size:  $4$
\item Epochs: $200$
\item Optimizer: The Adam algorithm\cite{kingma2017adammethodstochasticoptimization} with learning rate $0.001$, $\beta_1$ $0.9$, $\beta_2$ $0.999$.
\item Loss Function: Binary Crossentropy
\end{itemize}
After semantic segmentation, the vision system begins contour detection.

In contour detection, a border following algorithm\cite{Suzuki1985TopologicalSA} is used to generate boundaries between different regions generated by the semantic segmentation system. The contour of each region is used to generate a center point of the region. This position, if the region is pickable, is a potential target, but it must first be transformed from a pixel position into the coordinate frame of the robotic arm.

Once the location in the image frame is detected, we convert it to a coordinate for the motion planner in three steps. First, we use the camera intrinsic matrix, based on the field-of-view of the camera, and back projection to convert the 2D pixel coordinate into 3D space. Second, we transform the point from the coordinate frame of the camera depth sensor to the coordinate frame of the arm's base link. A Unified Robotics Description Format file defines the transform between the coordinate frames of the different objects. Third, we discard the calculated height: the z-axis distance. Instead, we set the height to be slightly higher than the robot base. The coordinate will move the robot above the target location rather than to it.  Later, during the control stage, the height is set to the infrared depth sensor distance. Because the end effector is pointed perpendicular to the ground, the sensor distance is the height off the ground. With a target coordinate in the robot base coordinate frame, the sample control system is ready to run.

\subsection*{Sample Control System}
The sample control system samples the soil at a position provided by the computer vision system.

The arm is controlled by a combination of a PID controller\cite{pid_controller} and motion planner. The motion planner, implemented in ROS MoveIt, uses the RRT* algorithm\cite{rrtstar} to plan motion between individual points and the PID controller creates a feedback loop allowing the arm to incrementally move closer to a given location.

Because some detected points may be unreachable, too far away for the arm, the computer vision system provides several candidate positions. The sample control system sends the coordinate corresponding to the largest approachable region, if any are close enough, to the motion planner.

\begin{figure}
  \centering
  \includegraphics[width=85mm]{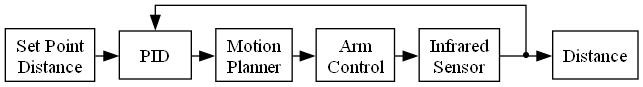}
  \caption{The distance detected by the infrared sensor is used as feedback.}
  \label{fig:pid-diagram1} 
\end{figure}

The motion planner finds a path for the arm that will allow it to reach the soil sample without colliding with itself or statically configured obstacles, like the platform it's mounted on. The arm first moves directly above the sample location. Because there is little risk of obstacles---the end effector is not near the ground yet---this is done without any sensor feedback. However, when the arm begins to dive towards the dirt, it switches to a feedback controlled mode. Because the infrared sensor is unreliable, the sample control system uses the motion planner iteratively. Feedback is managed by the PID controller, shown in Figure \ref{fig:pid-diagram1}. It helps the motion planner converge to the soil location.

As the arm dives, the depth sensor allows it to estimate its own position. Using the difference between measured positions and the setpoint, the arm changes its movement. Feedback is necessary because a dead reckoning system is imperfect, there would be drift, and the depth sensor used to measure distance from the ground varies in accuracy as the distance to the ground changes.

Although we call this a PID controller, it is proportional only. In experiments, as will be demonstrated, the system works. We refer to our controller as a PID controller to emphasize that it can be analyzed as a PID controller.

The controller proportionality constant is 0.9. We generate this constant empirically. Unlike a controller with an integral component, it suffers from proportional droop and slows down as it approaches the set point. This design was chosen to simplify the implementation. The set point is empirically measured to be 27cm off the ground: slightly shorter than the distance from the depth sensor to the tip of the gripper. We measure this by lowering the arm to the preferred pick height, waiting for the infrared sensor to steady, and recording the distance. This calibration is done once. While the control system runs, the arm repeats a loop of moving the gripper down by a distance equal to the difference between the desired position, the set point, and the depth sensor estimated position. This difference is weighted by the proportionality constant.

The failure handling mechanism is the last part of the control system. Physical systems can fail in more ways than virtual systems. Transferring from sim to real is a well studied problem. For example, Georgia Tech's robot terrarium is a test bed deliberately designed to provide a testing interface, similar to a simulator, for code meant to run on physical robots \cite{robot_tank}. This is necessary because what works in simulation---or theory---is often hard to translate to a physical system; there may be unexpected noise or faults. Our system is no exception. Our arm, can fail to see, approach, pick, or hold samples. To account for this and make our system more robust, we use a state machine.

\begin{figure}
  \centering
  \includegraphics[width=85mm]{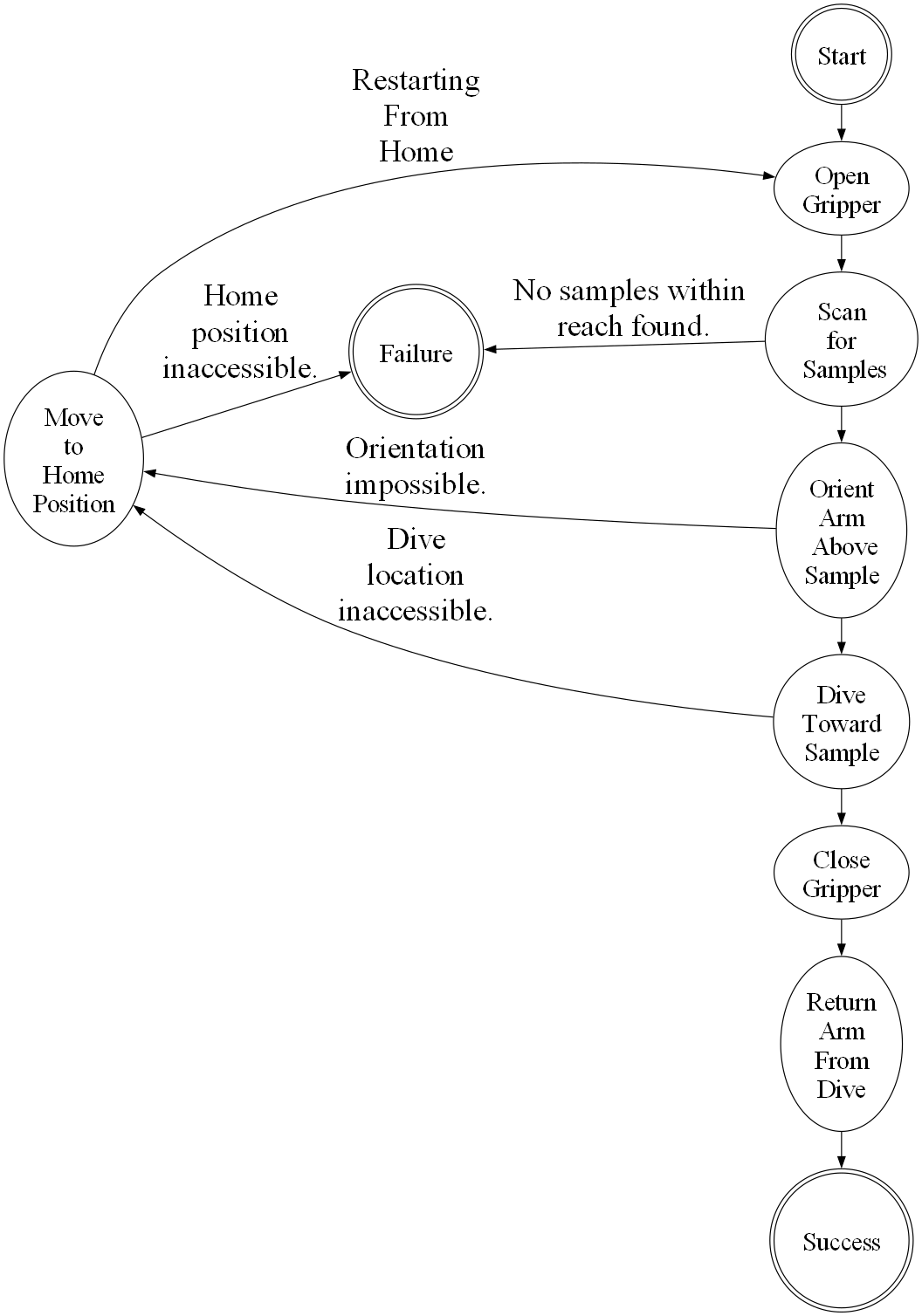}
  \caption{For some stages, the robot can retry on failure.}
  \label{fig:state-machine1} 
\end{figure}

The state machine handles failures by creating a series of recovery behaviors, similar to a behavior tree, but because the amount of states is low and we only need one control system per machine, we use a finite state machine.\cite{trees}. When the soil pick begins, the machine progresses from the start state, at the top right of Figure \ref{fig:state-machine1}, toward the bottom success state. If the system successfully transitions through the pick states, it will report success. Otherwise, it will report failure. Failure causes include a failure to detect any samples or an inability to reach any samples. For some of the failures, as noted in Figure \ref{fig:state-machine1}, the machine will attempt to move to an earlier state. This could loop forever, but in practice repeated recoverable failures eventually lead to an unrecoverable failure or success. In addition, another process can interrupt the system based on other criteria: maximum attempt time elapsed, user interrupt, or low battery state.

\section{EXPERIMENTS} \label{section:experiment}

In our experiments, we test the prototype pick success rate with a soil bed. In addition, we test the computer vision system in isolation and conduct an ablation study to measure the impact of auxiliary task training and data augmentation on the computer vision model.

\subsection*{Arm Performance Experiment}

To test the robustness of the arm, we run two experiments: one with rocks for the arm to avoid and one without any unpickable obstacles. Each experiment runs for 30 trials. In each trial, the robot starts from the same position, we begin the pick sequence, wait for the robot to complete a pick, and record the results. We record whether or not the soil collection was successful. If not, we record the cause of failure.

The experimental apparatus includes a mounting point for the arm and a soil bed for testing. The arm is mounted to an adjustable height machine table. The height is deliberately set to be the same height as a mounting point on a common robotics platform: a Clearpath Husky. The soil bed is a 40 by 40 inch bed that is 6 inches deep. It contains a loose collection of top soil and crushed rock, similar to Figure \ref{fig:hetero-material}.

\subsection*{Computer Vision Experiment}

To determine the effectiveness of our computer vision model, we test it in isolation. We measure the accuracy, precision, and Intersection over Union (IoU) of the model on the testing and validation datasets.

\subsection*{Ablation Experiment}

Because our dataset only contained 150 images, we used data augmentation and auxiliary task training to improve our model. To determine if either of these techniques helped, we conduct an ablation study. To determine if data augmentation improved the model, we train the model with and without data augmentation and then compare the results. We run another ablation study, with and without auxiliary task training, to measure its impact. For each ablation study, we record the accuracy, precision, and IoU on the testing and validation datasets.

\section{RESULTS} \label{section:results}

In our results, our prototype successfully collects dirt without rock obstacles (homogeneous terrain) and with rock obstacles (heterogeneous terrain). The prototype success rate is 6\% higher with rock obstacles. This is surprising because the computer vision system prevents the arm from hitting obstacles in all trials for both experiments; no failures are caused by misidentifying an obstacle. Instead, the extra 6\% failures are caused by failure to position the arm at the proper depth.

\subsection*{Arm Performance Experiments}

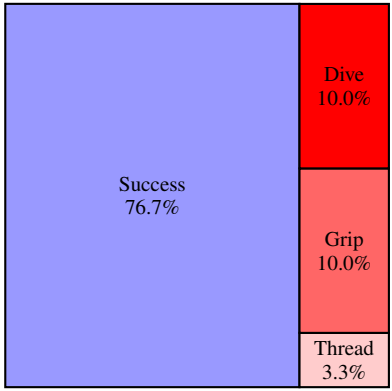
\begin{figure}
    \centering
    \begin{tikzpicture}[font=\footnotesize, scale=0.85]
      \pie[square,  
        color={blue!40, red!20, red!60, red!100}, 
        text=inside,
      ]{76.7/Success, 3.3/Thread, 10.0/Grip, 10.0/Dive}
    \end {tikzpicture}
    \caption{The results of the experiment on the homogeneous terrain: dirt.}
    \label{fig:dirt-results}
\end{figure}

\begin{figure}
    \centering
    \begin{tikzpicture}[font=\footnotesize, scale=0.85]
      \pie[square,
        color={blue!40, red!20, red!60},     
        text=inside,
      ]{83.3/Success, 10.0/Thread, 6.7/Grip}
    \end {tikzpicture}
    \caption{The results of the experiment on the heterogeneous terrain: dirt with rocks.}
    \label{fig:rock-results}
\end{figure}
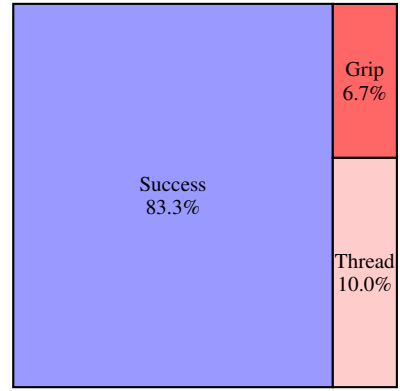

In the arm performance experiment, we discovered that the arm picker performed worse without rocks--6\% more failures--than with them. This is shown in Table \ref{tab:arm-results1}. We identified three failure types: failure to dive, failure to grip, and deadlock:
\begin{itemize}
\item In a failure to dive, the arm would position itself over dirt correctly, but then the arm would either submerge itself too deep within the material to close properly or not deep enough to pick anything. This was despite the PID controller and the depth sensor.
\item In a failure to grip, the arm would position itself over dirt correctly, but fail to hold the sample during the return. Instead, the sample fell out of the gripper.
\item In a deadlock failure, the arm would proceed to plan a trajectory, but the software hung while the arm was executing the motion plan.
\end{itemize}
Figure \ref{fig:dirt-results} and Figure \ref{fig:rock-results} summarize the results without and with rock obstacles, respectively.

\begin{table}[h!]
  \centering
    \caption{The confidence intervals of success rate on homogeneous and heterogeneous surfaces.}
    \label{tab:arm-results1}
    \begin{tabular}{l|r}
      $\textbf{Experiment}$ & $\textbf{Wilson Score}$ \\
      \hline
      Without Rocks & $73.6 \pm 14.6\%$ \\
      With Rocks & $79.6 \pm 13.1\%$ \\
    \end{tabular}
\end{table}

The prototype performance with and without rock obstacles is similar, but---to the computer vision system's credit---the arm never fails from collision with an obstacle. Instead, the arm dives too deep and fails to grip properly. One explanation for over diving is that the rocks are reflecting the depth sensor's infrared light more reliably than the dirt, but this should only cause the arm to dive too shallow.

The failures to grip are an opportunity to improve the system. Currently, the system cannot detect whether it has succeeded or not. If the system could detect this, then it could reattempt after a failure to grip. This would improve the success rate for both systems up to 6.7\%.

\subsection*{Computer Vision Experiment}
We trained the model for 200 epochs. We tracked accuracy, precision, IoU, and recall for the training, validation, and test sets. We achieved accuracy of 95.5\% for training and 94.8\% on the test set. A summary of all metrics and results are shown in Table \ref{tab:cv-results-stat}.

\begin{table}[h!]
  \centering
  \caption{Our results for the computer vision model.}
  \label{tab:cv-results-stat}
  \begin{tabular}{l|c|c|c|c}
    & $\textbf{Accuracy}$ & $\textbf{Precision}$ & $\textbf{IoU}$ & $\textbf{Recall}$ \\
    \hline
    $\textbf{Validation}$ & $95.76\%$ & $98.80\%$ & $92.36\%$ & $93.08\%$ \\
    $\textbf{Test}$ & $94.49\%$ & $95.67\%$ & $90.11\%$ & $87.83\%$\\
  \end{tabular}
\end{table}

The combination of U-Net\cite{unet1} and the several thousand training images, generated by data augmentation of the 150 images we collected, was cheap to produce but---as shown in the results---accurate enough for our prototype to collect soil.

\subsection*{Ablation Experiment}
For the ablation study, we tested the model without data augmentation and without auxiliary task training. The results without data augmentation are shown in Table \ref{tab:cv-ablation1}. The results without auxiliary task learning are shown in Table \ref{tab:cv-ablation2}. Without data augmentation, the model performed worse by all metrics except recall. Recall was slightly better without data augmentation. Without auxiliary task training, there was little to no change. These results suggest that data augmentation was crucial, but auxiliary task training had little to no effect. Auxiliary learning of a different task may be more effective.

\begin{table}[h!]
  \centering
  \caption{Performance Without Data Augmentation}
  \label{tab:cv-ablation1}
  \begin{tabular}{l|c|c|c|c}
    & $\textbf{Accuracy}$ & $\textbf{Precision}$ & $\textbf{IoU}$ & $\textbf{Recall}$ \\
    \hline
    $\textbf{Validation}$ & $47.8\%$ & $40.3\%$ & $30.2\%$ & $100.0\%$ \\
    $\textbf{Test}$ & $41.4\%$ & $33.9\%$ & $25.4\%$  & $100.0\%$ \\
    $\textbf{Validation} \Delta$ & ${ -48.01\%}$ & ${ -58.43\%}$ & ${ -62.19\%}$ & $+6.88\%$ \\    
    $\textbf{Test} \Delta$ & ${ -52.05\%}$ & ${ -61.8\%}$ & ${ -64.71\%}$ & $+12.14\%$ \\
  \end{tabular}
  \caption*{Besides recall, the performance without data augmentation is worse than with data augmentation.}
\end{table}

\begin{table}[h!]
  \centering
  \caption{Performance Without Auxiliary Task Training}
  \label{tab:cv-ablation2}
  \begin{tabular}{l|c|c|c|c}
    & $\textbf{Accuracy} \Delta$ & $\textbf{Precision} \Delta$ & $\textbf{IoU} \Delta$ & $\textbf{Recall} \Delta$ \\
    \hline
    $\textbf{Validation}$ & $95.3\%$ & $96.7\%$ & $91.1\%$ & $90.5\%$ \\
    $\textbf{Test}$ & $96.2\%$ & $96.0\%$ & $91.0\%$ & $89.1\%$\\
    $\textbf{Validation} \Delta$ & ${ -0.51\%}$ & ${ -2.10\%}$ & ${ -1.26\%}$ & ${ -2.56\%}$ \\
    $\textbf{Test} \Delta$ & $+1.71\%$ & $+0.28\%$ & $+0.9\%$ & $+1.26\%$ \\
  \end{tabular}
  \caption*{The performance without auxiliary task training is nearly the same as with auxiliary task training.}
\end{table}

\section{CONCLUSION} \label{section:conclusion}

In this paper, we designed, implemented, and tested a solution to the amorphous heterogeneous terrain sampling problem. We built a robotic arm control system capable of picking soil in an outdoor environment with applications to robotic agriculture, surveying, and hazard detection. In our results, our system collected soil from homogeneous terrain as easily as from heterogeneous terrain. The arm never attempted to pick up unpickable material. Our approach, unlike existing soil core samplers, uses an end effector capable of a variety of tasks. This makes it suitable to robots required to sample soil in addition to other tasks.

Future work may focus on two limitations: the lack of outdoor testing and the rigidness of the control system. Although the prototype was trained with outdoor data and could sample soil in preliminary outdoor tests, thorough experimental testing outside of the test apparatus is required. The control system is satisfactory for testing in a controlled environment, but an outdoor environment may require more recovery handling than our finite state machine can be programmed for. In addition, future work could focus on the effectiveness of different surrogate tasks for semantic segmentation or adapt computer vision research mentioned in Section \ref{section:related}, like the deep learning by Milioto, Lottes, and Stachniss\cite{8460962}.

\balance

\bibliographystyle{ieeetr}
\bibliography{sources}

\end{document}